%% file: main.tex
\documentclass[10pt,twocolumn,letterpaper]{article}

\usepackage[pagenumbers]{cvpr} 

\definecolor{cvprblue}{rgb}{0.21,0.49,0.74}
\usepackage[pagebackref,breaklinks,colorlinks,allcolors=cvprblue]{hyperref}
\usepackage[dvipsnames]{xcolor}
\usepackage{tikz}
\usepackage{pgfplots}
\usepackage{cancel}
\usepackage{fontawesome}
\usepackage{listings}
\usepackage[normalem]{ulem}
\usepackage{csquotes}
\usepackage{comment}
\usepackage{graphicx} 

\usepackage{booktabs}
\usepackage{multirow}
\usepackage{colortbl}
\usepackage{graphicx}

\usepackage[table]{xcolor}
\usepackage{colortbl}
\definecolor{lightgreen}{RGB}{200, 255, 200}
\usepackage{enumitem}

\usepackage[T1]{fontenc}
\usepackage{textcomp}
\usepackage{listings}
\usepackage{inconsolata}   
\usepackage{upquote}       
\usepackage{xcolor}

\definecolor{jsonkey}{HTML}{1A237E}
\definecolor{jsonstring}{HTML}{0B8043}
\definecolor{jsonnumber}{HTML}{B71C1C}
\definecolor{jsonbool}{HTML}{B45309}
\definecolor{listingbg}{HTML}{F8F9FA}
\definecolor{listingrule}{HTML}{E0E3E7}

\lstdefinelanguage{json}{
  basicstyle=\ttfamily\footnotesize,
  showstringspaces=false,
  columns=fullflexible,
  keepspaces=true,
  breaklines=true,
  breakatwhitespace=false,
  morestring=[b]",
  stringstyle=\color{jsonstring},
  literate=
   *{true}{{{\color{jsonbool}true}}}{4}
    {false}{{{\color{jsonbool}false}}}{5}
    {null}{{{\color{jsonbool}null}}}{4}
    {0}{{{\color{jsonnumber}0}}}{1}
    {1}{{{\color{jsonnumber}1}}}{1}
    {2}{{{\color{jsonnumber}2}}}{1}
    {3}{{{\color{jsonnumber}3}}}{1}
    {4}{{{\color{jsonnumber}4}}}{1}
    {5}{{{\color{jsonnumber}5}}}{1}
    {6}{{{\color{jsonnumber}6}}}{1}
    {7}{{{\color{jsonnumber}7}}}{1}
    {8}{{{\color{jsonnumber}8}}}{1}
    {9}{{{\color{jsonnumber}9}}}{1},
}

\lstdefinestyle{jsonstyle}{
  language=json,
  backgroundcolor=\color{listingbg},
  frame=single,
  framerule=0.6pt,
  rulecolor=\color{listingrule},
  xleftmargin=0pt,
  aboveskip=6pt,
  belowskip=6pt,
  numbers=none,
  upquote=true,
  postbreak=\mbox{\textcolor{gray}{$\hookrightarrow$}\space}
}

\lstdefinestyle{promptstyle}{
  basicstyle=\ttfamily\footnotesize,
  backgroundcolor=\color{listingbg},
  frame=single,
  framerule=0.6pt,
  rulecolor=\color{listingrule},
  columns=fullflexible,
  keepspaces=true,
  breaklines=true,
  breakatwhitespace=false,
  numbers=none,
  upquote=true,
  aboveskip=6pt,
  belowskip=6pt,
  postbreak=\mbox{\textcolor{gray}{$\hookrightarrow$}\space}
}

\title{DamageScope: Vision–Language Retrieval at Scale for Disaster Damage Assessment from Satellite Imagery}

\author{Ravi K. Rajendran$^{*}$, Biplob Debnath$^{*}$, Murugan Sankaradas$^{\dagger}$ and Srimat T. Chakradhar\\
NEC Laboratories America, Princeton, NJ, USA\\
{\tt\small Email: \{rarajendran,biplob,murugs,chak\}@nec-labs.com}
}

\begin{document}
\maketitle
\renewcommand{\thefootnote}{\fnsymbol{footnote}}
\footnotetext[1]{\ Equal contribution.}
\footnotetext[2]{\ Work performed while at NEC Laboratories America.}

\begin{abstract}
    \input{src/abstract}
\end{abstract}

\input{src/introduction}
\input{src/method}
\input{src/evaluation}
\input{src/related-work}
\input{src/conclusion}

{
    \small
    \bibliographystyle{ieeenat_fullname}
    \bibliography{reference}
}

\input{src/appendix}

\end{document}

%% file: src/abstract.tex
Timely and accurate assessment of property damage is critical following natural disasters. Traditional on-site inspections are labor-intensive, costly, and often pose safety risks. Advances in satellite imagery and vision-language models (VLMs) enable scalable remote damage assessment; however, integrating VLMs into large-scale Earth observation pipelines presents challenges in computational efficiency, data organization, and information retrieval. To address these challenges, we present DamageScope, a retrieval-augmented framework that combines satellite imagery with Vision-Language Models (VLMs) and Large Language Models (LLMs) to automate property damage analysis. Built on a Retrieval-Augmented Generation (RAG) framework, DamageScope extracts structured visual representations from satellite imagery to support interactive natural language queries for damage assessment.
To address scalability, we introduce a novel multi-vector embedding-based clustering algorithm that outperforms traditional single-vector embedding approaches while reducing indexing time by up to $14\times$. Furthermore, a dual-store data architecture minimizes LLM API calls, reducing both operational cost and response latency by up to approximately $3\times$. By effectively balancing scalability and operational efficiency, DamageScope provides a robust and practical solution for real-world damage assessment tasks.

%% file: src/introduction.tex
\section{Introduction}
Accurate and timely assessments of property damage play a critical role in disaster recovery, particularly in the aftermath of natural disasters such as hurricanes, earthquakes, or floods. These evaluations are essential for determining the extent of damage, expediting insurance claims, and facilitating the restoration process for affected individuals and communities. However, traditional methods of damage assessment primarily rely on on-site inspections conducted by field adjusters. While this approach has been the standard practice for years, it comes with several challenges and limitations. On-site inspections are often costly due to the resources required, leading to delays in processing claims and increased operational expenses. Additionally, they can be time-consuming, as adjusters must physically visit each affected property to document and evaluate the damage. This process can be further delayed when access to certain areas is restricted due to infrastructure damage or hazardous conditions caused by the disaster. Given these limitations, there is a growing need for more efficient and safer alternatives to traditional damage evaluation methods.

\begin{figure}[t!]
    \centering
    \includegraphics[width=\linewidth]{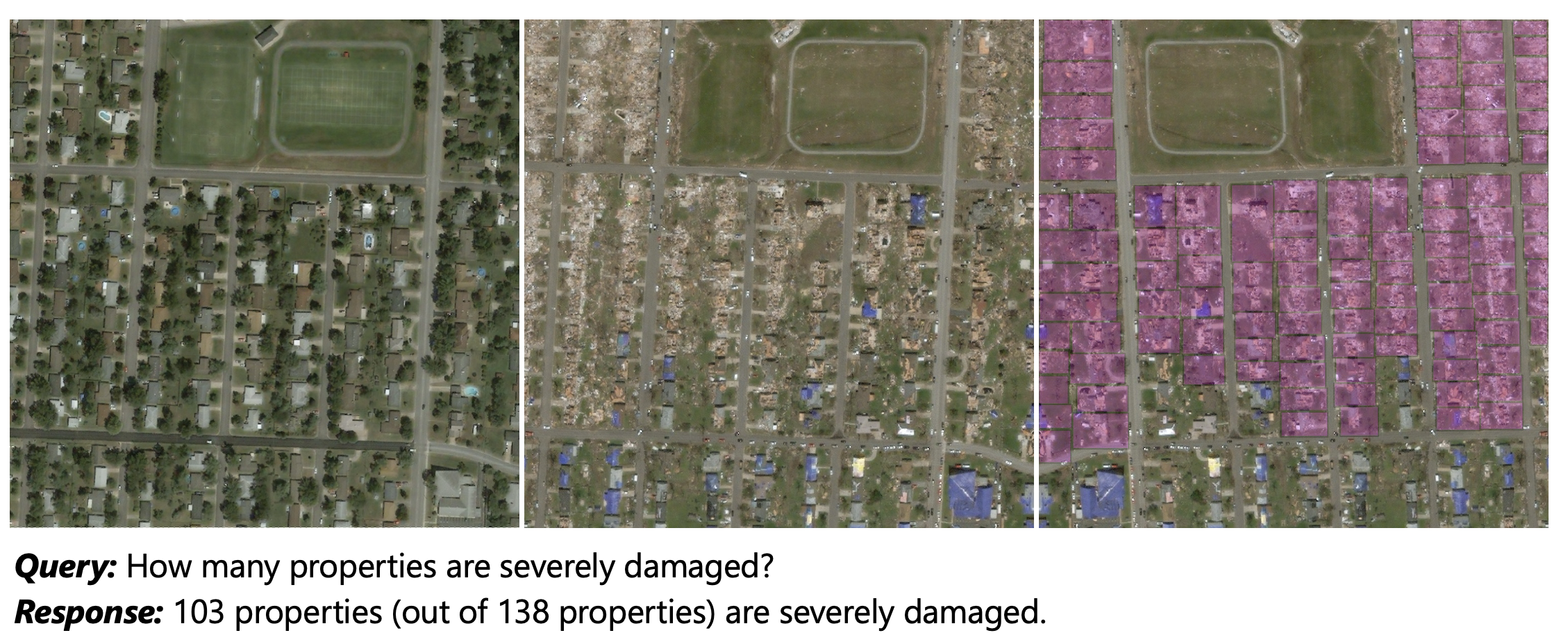}
    \caption{DamageScope's response to the query, \enquote{List all the properties that are severely damaged.} The figure shows the pre- (left), post-disaster image (middle), and annotated response highlighting the damaged properties (right).}
    \label{fig:teaser-figure}
\end{figure}

Recent advancements in satellite imaging and artificial intelligence (AI) are transforming this damage assessment process. Satellite imagery enables insurers to conduct rapid damage assessments over extensive areas without the need for physical on-site visits. High-resolution satellite images allow insurers to quickly gauge the extent of damage across different types of properties - be it structural, roof, or environmental damage while minimizing human intervention~\cite{brunner2010, wang_2023_dual_path}. The integration of AI with satellite data further enhances this process by automating the analysis of imagery to detect and quantify damage. Machine learning models, particularly deep learning techniques such as convolutional neural networks (CNNs) and transformers, have demonstrated high accuracy in damage classification tasks \cite{ doshi2018satelliteimagerydisasterinsights, hao_2020_attention_based}. More recently, vision-language models (VLMs)  that fuse satellite images with metadata have been used to improve assessment precision. These advancements enable insurers to process claims more quickly, make well-informed decisions with minimal delays, and ultimately improve the overall customer experience.

In this paper, we introduce \textit{DamageScope}, a system designed to enhance property damage analysis by integrating satellite imagery, AI models, and large language models (LLMs). DamageScope is built on the retrieval-augmented generation (RAG) framework~\cite{rag}, enabling efficient retrieval and reasoning over damage data. Damage analysis begins with a satellite image as the primary input. The system uses property metadata, such as geographical boundaries and property records, to isolate individual properties within the image. Once each property image is extracted, it is processed by a vision-language model (VLM), such as GPT-4o~\cite{openai2025}, which is capable of analyzing both visual data and corresponding contextual information. The VLM analyzes an image to extract various damage-related metadata, including structural integrity, roof damage, water accumulation, habitability status, and overall damage score. This extracted information is then saved in data stores for later querying.  

DamageScope allows insurers to interact with the system using natural language queries, facilitating seamless access to critical insights. Given a satellite image, now users can ask detailed questions about damage severity, affected regions, or specific property conditions. The system retrieves relevant information from the data store and provides it as context to an LLM \cite{openai2025} along with the query, which then generates a comprehensive, context-aware response.  For example, Figure~\ref{fig:teaser-figure} illustrates the DamageScope’s output for a query identifying \enquote{severely damaged properties}.\\

\vspace{-6pt}
\noindent\textbf{Challenges.} Deploying the DamageScope system presents several challenges that must be addressed to ensure efficiency, scalability, and accuracy in property damage assessment. \\

\vspace{-6pt}
\noindent\textbf{Challenge \#1: Computational Efficiency in Damage Metric Generation.} 
Extracting and analyzing damage metrics from property images is a computationally intensive task, particularly in large-scale disaster scenarios involving thousands of properties. Processing each image individually using VLMs is prohibitively slow and expensive. To address this, DamageScope employs a clustering-based strategy that groups properties with similar characteristics, thereby reducing redundant computation.

While traditional clustering methods typically rely on single-vector embeddings, we utilize \textit{multi-vector embeddings}~\cite{colbert, colpali, scheerer2025warpefficientenginemultivector}, which have shown superior performance in information retrieval tasks~\cite{xtr, scheerer2025warpefficientenginemultivector, plaid}. To quantify similarity between two multi-vector representations, the $\text{MaxSim}$ function~\cite{colbert, colpali, xtr} is widely used, as it identifies the maximum token-level similarity across the two sets of vectors. However, $\text{MaxSim}$ is inherently asymmetric, i.e., $\text{MaxSim}(A, B) \neq \text{MaxSim}(B, A)$, which limits its applicability in clustering algorithms that require symmetric distance measures.

To overcome this limitation, we propose a \textit{symmetric variant} of the $\text{MaxSim}$ function. By averaging the directional $\text{MaxSim}$ scores between two sets of vectors, the resulting $\text{MaxSim}_{\text{sym}}$ yields a reciprocal similarity measure, enabling consistent distance computations essential for clustering.  

\vspace{0.25cm}
\noindent\textbf{Challenge \#2: Efficient Organization and Retrieval of Damage Information.}  Once damage metrics are generated, it is essential to structure the data in a way that facilitates efficient retrieval. Storing all damage information within a single store can lead to slow and inaccurate retrieval processes. To address this challenge, DamageScope employs a multi-store data organization strategy, distributing the damage data across multiple data stores, thus optimizing retrieval time and accuracy.

\vspace{0.25cm}
\noindent\textbf{Challenge \#3: Optimized Query Processing with Large Language Models.} DamageScope leverages LLMs to generate concise, human-like responses to user queries. However, it faces two main challenges: (1) LLMs are constrained by context size limitations, restricting the amount of input they can process at any given time, and (2) the cost of using the LLM API can be high~\cite{AREFEEN2024100065}. To overcome these challenges, DamageScope employs a strategic data filtering mechanism leveraging LLMs. Rather than feeding the entire data to be processed, the DamageScope uses LLM-driven planning to selectively retrieve the most relevant information from the data store to form the context for response generation, optimizing both performance and cost-efficiency. 

\begin{figure*}[t]
\centering
\includegraphics[width=0.90\textwidth]{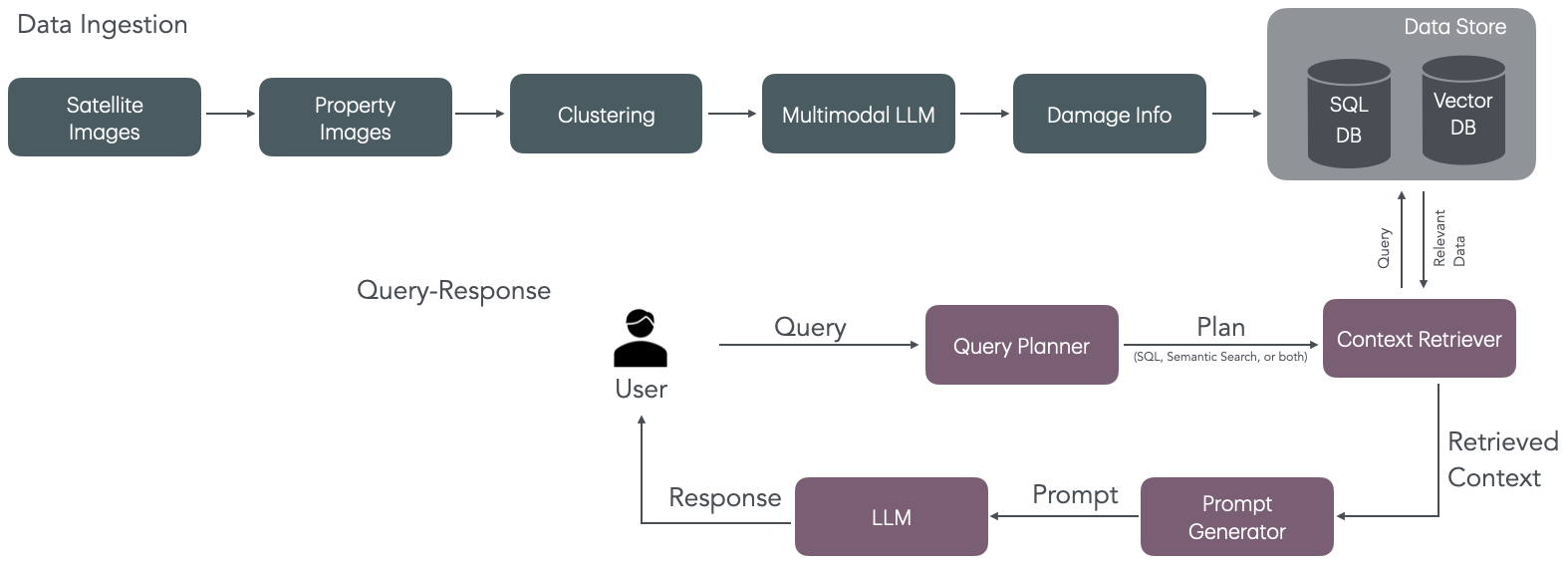}
\caption{DamageScope workflow. The ingestion pipeline (top) converts satellite imagery into structured and semantic representations stored in a dual data store. The query-response pipeline (bottom) retrieves relevant evidence and composes an LLM-grounded response.}
\label{fig:damagescope}
\end{figure*}

\vspace{0.2cm}

By addressing these challenges - reducing computational costs through clustering, optimizing data retrieval through multiple stores, and improving query efficiency through context-aware LLM integration, DamageScope achieves a balance between accuracy, scalability, and efficiency. In summary, we make the following contributions:

\begin{itemize} [left=0pt]
    \item We introduce \textit{DamageScope} system that integrates satellite imagery, Vision Language Models (VLMs), and Large Language Models (LLMs) to perform property damage assessment. It supports natural language querying, allowing users to retrieve damage insights.

    \item We present a clustering algorithm based on multi-vector embeddings to improve the computational efficiency of property-level damage information generation. Evaluation on the xBD~\cite{xu_2019_building_damage_detection} benchmark shows that our multi-vector clustering approach outperforms its single-vector counterpart, achieving gains of up to 0.20 in Normalized Mutual Information (NMI)~\cite{vinh2010information}, 0.26 in Fowlkes-Mallows Score (FMS)~\cite{fowlkes1983method}, and 0.26 in Adjusted Rand Index (ARI)~\cite{hubert1985comparing}. Additionally, it reduces overall indexing time up to a factor of $14\times$.

    \item Our evaluation shows that the question-answering capability of DamageScope is approximately $3\times$ more cost-efficient and $2.9\times$ faster compared to baseline approaches. This performance gain is achieved through a dual-store data organization that distributes the data across multiple data stores, optimizing retrieval latency and resource efficiency.
\end{itemize}

%% file: src/method.tex
\section{DamageScope System}
\label{sec:damagescope-system}
DamageScope builds on the Retrieval-Augmented Generation (RAG) paradigm, integrating multimodal perception and language-based reasoning to support scalable, interpretable damage assessment. The system is designed around the real-world requirements of \emph{speed}, \emph{scalability}, and \emph{traceability}. It operates in two stages:
(1)~data ingestion, which converts raw satellite imagery into structured representations, and (2)~question answering, which retrieves relevant evidence and synthesizes natural-language responses. An overview of the DamageScope system workflow is shown in Figure~\ref{fig:damagescope}.

\subsection{Data Ingestion: Extracting Structured Damage Information}
Traditional satellite pipelines use pixel-level models~\cite{doshi2018satelliteimagerydisasterinsights, xu_2019_building_damage_detection}, producing narrow outputs that are hard to reuse for reasoning or retrieval. DamageScope instead generates a semantically rich representation for each property, combining categorical attributes (e.g., roof integrity, structural damage) with natural-language rationales grounded in the imagery.

In DamageScope, analysis begins with a satellite image and property metadata, which define geographic boundaries and allow extraction of individual property images. Each property crop is then processed by a vision-language model (VLM) prompted to return a JSON summary with categorical predictions and natural-language reasoning. 

A sample output is shown in Listing~\ref{lst:sample-damage-info}. Such structured outputs enable both factual querying (e.g., damage\_level) and semantic retrieval over reasoning. However, naively invoking a VLM on every building introduces significant computational overhead: a single GPT-4o~\cite{openai2025} inference takes \(\approx 8.4\) seconds. For a mid-sized city with ~10,000 building footprints, this translates to \(\sim\) 23 hours of sequential processing, which is far too slow for rapid post-disaster assessment. Although parallelization can reduce wall-clock latency, it does not address the underlying scalability and cost constraints inherent to a \enquote{VLM-per-property} pipeline. The next section describes how DamageScope addresses this bottleneck through clustering.

\begin{lstlisting} [style=jsonstyle,caption={Example structured damage summary.},label={lst:sample-damage-info}]
{
  "property_id": "A12-487",
  "roof_damage": {
    "prediction": true,
    "reason": "Significant portions of the roof shingles are missing."
  },
  "damage_level": {
    "prediction": 3,
    "reason": "Major roof loss with likely internal damage."
  }
}
\end{lstlisting}

\begin{figure*}
    \centering
    \includegraphics[width=\linewidth,clip,trim=0cm 11.5cm 0.8cm 2.9cm]{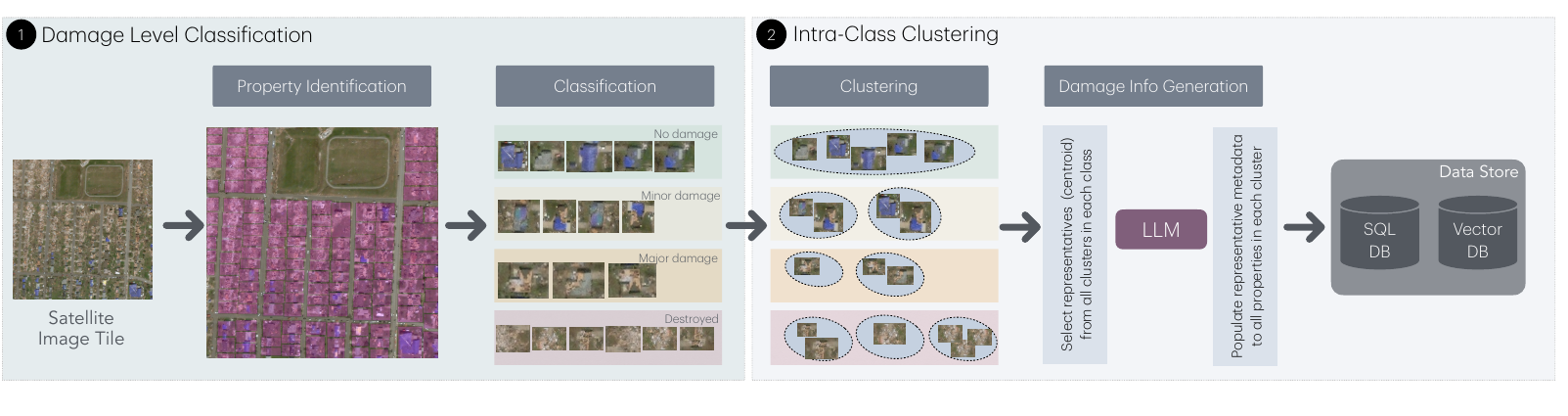}
    \caption{Damage information generation workflow using clustering. First, images are classified into four groups. Next, the \enquote{No Damage} group contains a single cluster, while the other groups are further clustered using multi-vector embeddings. Damage information is then generated for each cluster centroid using a LLM. Finally, damage information is stored in the data store.}
    \label{fig:data-ingestion}
\end{figure*}

\subsubsection{Clustering for Scalable Ingestion}
Disaster damage exhibits spatial correlation: nearby properties often share similar destruction patterns. To avoid redundant VLM inference, DamageScope performs clustering and analyzes only representative items from each cluster. 

To form clusters, DamageScope uses a two-step process as shown in Figure~\ref{fig:data-ingestion}. Each property image is classified into one of four categories: 0 (No Damage), 1 (Minor), 2 (Major), or 3 (Destroyed), using labeled data such as the xBD dataset \cite{gupta_2019_xbddataset}. Properties labeled \enquote{No Damage} require no further analysis. For all other classes, DamageScope performs clustering within each category to capture finer-grained, structurally similar damage patterns.

This two-step pipeline, consisting of (1) coarse classification followed by (2) intra-class clustering, improves both efficiency and representational quality. The initial classifier reduces heterogeneity by partitioning images into coarse severity groups, creating more homogeneous subsets for clustering. Applying clustering only within non-zero classes then reveals coherent, fine-grained damage patterns that would be obscured in a single global clustering step. This design reduces noise, improves cluster stability, and avoids unnecessary computation for undamaged properties, enabling more interpretable and scalable analysis.

\subsubsection{Intra-Class Clustering}
\label{sec:intra-class-clustering}
\paragraph{Multi-Vector Embedding.} Single-vector embeddings compress spatial information into a single point, making them insufficient for capturing fine-grained cues such as shingle loss, debris distribution, or partial collapse. DamageScope therefore adopts multi-vector embeddings, which preserve token-level information across the image grid and have demonstrated strong performance in late-interaction retrieval models~\cite{colbert, colpali, plaid, scheerer2025warpefficientenginemultivector}.

A challenge arises because standard late-interaction similarity scoring functions, such as MaxSim~\cite{colbert}, are \emph{asymmetric}:
\begingroup
\setlength{\abovedisplayskip}{4pt}
\setlength{\belowdisplayskip}{4pt}
\[
\text{MaxSim}(A,B) \neq \text{MaxSim}(B,A)
\]
\endgroup
as the score averages over tokens from only one embedding set. While acceptable for ranking applications, this asymmetry makes MaxSim unsuitable as a \emph{distance function} for clustering. In general, clustering algorithms, whether hierarchical~\cite{UPGMA}, centroid based (e.g., k-means variants~\cite{kMeans}), or density based~\cite{ester1996}, require a \textit{symmetric} pairwise distance measure; otherwise, the resulting distance matrix becomes ill formed and produces inconsistent cluster assignments that depend on arbitrary ordering rather than true similarity.

To enable clustering with multi-vector representations, we introduce a symmetric extension of MaxSim.  
Given multi-vector embeddings \(A=\{a_i\}_{i=1}^m\) and \(B=\{b_j\}_{j=1}^n\), and a similarity function \(\texttt{sim}(\cdot,\cdot)\) (e.g., cosine similarity), we define:
\begingroup
\setlength{\abovedisplayskip}{4pt}
\setlength{\belowdisplayskip}{4pt}
\begin{align}
    \text{MaxSim}_{\text{sym}}&(A,B)= \tfrac{1}{2}\!\Bigg(\tfrac{1}{m}\!\sum_i \max_j \text{sim}(a_i,b_j) \nonumber \\
    &\quad + \tfrac{1}{n}\!\sum_j \max_i \text{sim}(b_j,a_i)\Bigg)
\end{align}
\endgroup

This formulation guarantees:
\begingroup
\setlength{\abovedisplayskip}{4pt}
\setlength{\belowdisplayskip}{4pt}
\[
\text{MaxSim}_{\text{sym}}(A,B)=\text{MaxSim}_{\text{sym}}(B,A)
\]
\endgroup
allowing us to define a valid distance metric
\begingroup
\setlength{\abovedisplayskip}{4pt}
\setlength{\belowdisplayskip}{4pt}
\[
d(A,B)=1-\text{MaxSim}_{\text{sym}}(A,B).
\]
\endgroup

Using this symmetric metric produces coherent, stable clusters that reflect mutual visual similarity between damaged properties. In practice, this improves cluster purity and provides a reliable basis for reducing redundant ingestion computations.

\paragraph{Clustering Algorithm.}
DamageScope clusters embedded properties and sends only one representative per cluster to the VLM. It uses hierarchical agglomerative clustering (HAC)~\cite{UPGMA} because it aligns with the structure of multi-vector embeddings. Damage patterns often form \emph{non-convex}, \emph{heterogeneous}, and \emph{uneven-sized} groups, conditions under which k-means~\cite{kMeans} fails due to its spherical-cluster assumption. Density-based methods such as DBSCAN~\cite{ester1996} also struggle with scale-sensitive thresholds in high-dimensional token spaces. HAC naturally supports our symmetric pairwise distance \(d(A,B)\) without requiring centroid computation and better preserves semantic neighborhoods. For a class with \(N_c\) samples, we set the number of clusters to
\[
K_c = \lfloor \sqrt{N_c} \rfloor,
\]
following heuristics from ColBERTv2~\cite{colbertv2}.

\subsubsection{Dual-Store Data Organization}
\label{sec:dual-store}
Ingestion produces two distinct information types: (i) \emph{structured attributes} (e.g., damage indicators, severity levels) and (ii) \emph{textual reasoning} describing visual evidence. Treating these uniformly is inefficient: relational databases excel at structured filtering but perform poorly on semantic search, while vector stores support embedding-based retrieval but are inefficient for grouping, aggregation, and large-scale filtering.

To address this, DamageScope uses a dual-store architecture that assigns each data type to the system best suited for it. Structured metadata is stored in SQL database (DB), enabling fast indexing and precise queries such as \textit{``Filter all properties with \texttt{damage\_level}~$\geq$~2''} or \textit{``List properties with structural damage''}. Textual reasoning is embedded and stored in a vector DB, supporting open-ended retrieval such as \textit{``Find properties with signs of roof shingle loss''} or \textit{``Retrieve all houses described as uninhabitable''}.

\begin{lstlisting}[style=jsonstyle,caption={Structured metadata stored in SQL DB.},label={lst:sample-damage-info-sql}]
{
  "property_id": "A12-487",
  "roof_damage": true,
  "damage_level": 3
}
\end{lstlisting}

The SQL database stores compact, schema-aligned records, as shown in Listing~\ref{lst:sample-damage-info-sql}. The vector database retains semantically rich reasoning information, as illustrated in Listing~\ref{lst:sample-damage-info-vectordb}. By routing each query to the appropriate store, DamageScope avoids unnecessary text processing and prevents large reasoning fields from inflating the LLM context window.

\begin{lstlisting}[style=jsonstyle,caption={Reasoning fields stored in the vector DB.},label={lst:sample-damage-info-vectordb}]
{
  "property_id": "A12-487",
  "roof_damage_reason":
    "Significant portions of the roof shingles are missing.",
  "damage-level_reason":
    "Major roof loss with likely internal damage"
}
\end{lstlisting}

\subsection{Question Answering: Retrieval Aware LLM Reasoning}
After ingestion, users query DamageScope in natural language (see the bottom part of Figure~\ref{fig:damagescope}). These queries leverage the two data types defined in the dual-store architecture described in Section~\ref{sec:dual-store}: structured attributes (e.g., \texttt{damage\_level}) and textual reasoning fields (e.g., \texttt{roof\_damage\_reason}). Sending all fields to an LLM is impractical due to context length limitations and cost constraints.

To manage this, DamageScope employs a retrieval aware query planner that determines which store, SQL or vector, is required for a given question. Metadata oriented queries (e.g., \textit{``List properties with \texttt{damage\_level}~$\geq$~2''}) are translated into SQL filters, while descriptive queries (e.g., \textit{``Find houses with missing shingles''}) trigger vector similarity search over reasoning embeddings. Questions that require both evidence types retrieve from both stores. DamageScope then assembles a compact prompt containing the user query and only the relevant structured fields and reasoning snippets (e.g., \texttt{roof\_damage=true} or ``Significant portions of the roof shingles are missing''). This ensures grounded responses while keeping token usage and latency low.

\subsection{Design Insights}
DamageScope’s efficiency arises from the interaction of its components rather than any single mechanism. Multi-vector clustering reduces ingestion cost by exploiting spatial correlation in damage patterns, allowing the system to process only a small set of representative properties. The dual store architecture streamlines retrieval by separating fast relational queries from semantic similarity search so that each query retrieves only the evidence it requires. The LLM based query planner coordinates these capabilities by interpreting user intent, issuing appropriate SQL and vector queries, and keeping LLM context small and responses traceable. Together, these choices reflect a broader principle: combining retrieval aware reasoning with domain specific data structuring enables large language models to operate efficiently and reliably in high stakes, non textual settings such as post disaster damage assessment.

%% file: src/evaluation.tex
\section{Experiments}
\label{sec:experiments}

In this section, we evaluate the quality of the clusters, the impact of clustering, data organization, and the LLM API usage cost optimization of the DamageScope system.

\newcommand{\greenup}{\textcolor{Green}{$\uparrow$}}
\newcommand{\greendown}{\textcolor{Green}{$\downarrow$}}
\newcommand{\redup}{\textcolor{Red}{$\uparrow$}}
\newcommand{\reddown}{\textcolor{Red}{$\downarrow$}}

\begin{table*}[t]
\caption{Clustering performance comparison between single and multi-vector embeddings using \textbf{RemoteCLIP} and \textbf{SigLIP} models. \cellcolor{lightgreen}Green cells indicate best per method and metric; \textbf{bold} indicates best overall per damage type.}
\label{tab:models_final_shaded}
\centering
\small
\begin{tabular}{ccccccccc}
\toprule
\multirow{2}{*}{\textbf{Image ID}} & \multirow{2}{*}{\textbf{Damage}} & \multirow{2}{*}{\textbf{Method}} &
\multicolumn{3}{c}{\textbf{RemoteCLIP}}  & 
\multicolumn{3}{c}{\textbf{SigLIP}} \\
\cmidrule(lr){4-6}\cmidrule(lr){7-9}
 &  & & \textbf{NMI} & \textbf{FMS} & \textbf{ARI} & \textbf{NMI} & \textbf{FMS} & \textbf{ARI} \\
\midrule

\multirow{2}{*}{31} & \multirow{2}{*}{Minor} 
& Single-vector + kMeans      & \cellcolor{lightgreen} 0.428 & \cellcolor{lightgreen} 0.283 & \cellcolor{lightgreen} 0.177 & 0.372 & 0.216 & 0.104 \\
& & Multi-vector + UPGMA       & \cellcolor{lightgreen}\textbf{0.561} & \cellcolor{lightgreen}\textbf{0.521} & \cellcolor{lightgreen}\textbf{0.416} & 0.337 & 0.338 & 0.096 \\
\midrule

\multirow{2}{*}{31} & \multirow{2}{*}{Major} 
& Single-vector + kMeans      & 0.507 & 0.522 & 0.288 & \cellcolor{lightgreen}\textbf{0.787} & \cellcolor{lightgreen}0.658 & \cellcolor{lightgreen}0.521 \\
& & Multi-vector + UPGMA       & 0.704 & 0.681 & \cellcolor{lightgreen}\textbf{0.544} & \cellcolor{lightgreen}0.724 & \cellcolor{lightgreen}\textbf{0.733} & 0.516 \\
\midrule

\multirow{2}{*}{31} & \multirow{2}{*}{Destroyed} 
& Single-vector + kMeans      & 0.543 & 0.366 & 0.225 & \cellcolor{lightgreen}0.544 & \cellcolor{lightgreen}0.399 & \cellcolor{lightgreen}0.239 \\
& & Multivector + UPGMA       & \cellcolor{lightgreen}\textbf{0.561} & \cellcolor{lightgreen}\textbf{0.567} & 0.365 & 0.529 & 0.526 & \cellcolor{lightgreen}\textbf{0.393} \\

\specialrule{\heavyrulewidth}{2pt}{1pt}
\specialrule{\heavyrulewidth}{0.5pt}{3pt}

\multirow{2}{*}{16} & \multirow{2}{*}{Minor} 
& Single-vector + kMeans      & 0.526 & \cellcolor{lightgreen}0.435 & \cellcolor{lightgreen}0.301 & \cellcolor{lightgreen}0.549 & 0.384 & 0.265 \\
& & Multi-vector + UPGMA       & \cellcolor{lightgreen}\textbf{0.694} & \cellcolor{lightgreen}\textbf{0.669} & \cellcolor{lightgreen}\textbf{0.565} & 0.376 & 0.333 & 0.114 \\
\midrule

\multirow{2}{*}{16} & \multirow{2}{*}{Major} 
& Single-vector + kMeans      & \cellcolor{lightgreen}0.573 & \cellcolor{lightgreen}0.473 & \cellcolor{lightgreen}0.331 & 0.493 & 0.413 & 0.265 \\
& & Multi-vector + UPGMA       & 0.519 & 0.514 & 0.271 & \cellcolor{lightgreen}\textbf{0.610} & \cellcolor{lightgreen}\textbf{0.575} & \cellcolor{lightgreen}\textbf{0.380} \\
\midrule

\multirow{2}{*}{16} & \multirow{2}{*}{Destroyed} 
& Single-vector + kMeans      & 0.422 & \cellcolor{lightgreen}0.272 & \cellcolor{lightgreen}0.132 & \cellcolor{lightgreen}0.432 & 0.257 & 0.126 \\
& & Multi-vector + UPGMA       & \cellcolor{lightgreen}\textbf{0.446} & 0.428 & 0.216 & 0.421 & \cellcolor{lightgreen}\textbf{0.477} & \cellcolor{lightgreen}\textbf{0.218} \\
\bottomrule

\end{tabular}
\end{table*}

\subsection{Implementation Details}
\textit{Data Ingestion:} The ingestion phase begins with data preprocessing, where satellite images are processed to extract property information rather than just building details. This preprocessing ensures that assessments account for the entire property, capturing a more comprehensive view of disaster impacts. Each property is identified and segmented for downstream analysis. Damage information is extracted using GPT-4o~\cite{openai2025}, which analyze and annotate the images for damage metrics (e.g., roof damage, water damage, etc.) and textual reasoning. 
To store the extracted data, we use dual-storage mechanisms to optimize query response time and accuracy. For damage metrics, we used SQLite~\cite{sqlite2020hipp}. For textual reasoning, we used ChromaDB~\cite{chromadb2023} for vector storage. This approach ensures that SQL queries and semantic searches are executed efficiently.

\noindent \textit{Question Answering (QA):} QA phase relies on the dual storage mechanism to optimize query response time and accuracy. Responses are generated using GPT-4o, which is used to make query plans, SQL queries, and other retrieval prompts. GPT-4o’s ability to reason about complex queries ensures that relevant information is fetched effectively from the data stores, thus minimizing unnecessary computations and enhances overall system performance.

To evaluate the effectiveness of DamageScope, we compare against a \textit{baseline} system that does not use clustering for ingestion, uses a single data store, and does not perform query-aware content selection during question answering.

\subsection{Dataset}
For our experiments, we utilize a publicly available disaster impact dataset, including satellite imagery and property damage assessments. Specifically, we used xBD dataset ~\cite{gupta_2019_xbddataset}, which provides pre- and post-disaster satellite images of properties affected by natural disasters such as hurricanes, tornadoes, etc., as well as proprietary customer data to enhance the robustness of our evaluation. xDB dataset focuses on structural damage but includes only building polygons, lacking information about the entire property. To address this limitation, we performed additional preprocessing and annotation to delineate complete property boundaries. These boundaries encompass not only the building but also surrounding features such as lawns, backyards, front yards, and driveways, providing a more holistic view of disaster impact.

\subsection{Results}
\subsubsection{Clustering Performance Evaluation}
Table~\ref{tab:models_final_shaded} presents a comparative analysis of clustering performance across different damage types and embedding strategies, using images from the Joplin tornado subset of the xBD dataset~\cite{xu_2019_building_damage_detection}. We evaluate the effectiveness of two vision-language models: RemoteCLIP~\cite{liu2024remoteclip}
(\texttt{chendelong/RemoteCLIP ViT-B-32}) and SigLIP~\cite{sigLIP} (\texttt{google/siglip-so400m- patch14-384}). Images are processed using the vision encoder, and we extract the output from the final transformer layer to obtain token-level image features. 

We use three standard clustering metrics: Normalized Mutual Information (NMI)~\cite{vinh2010information}, Fowlkes-Mallows Score (FMS)~\cite{fowlkes1983method}, and Adjusted Rand Index (ARI)~\cite{hubert1985comparing}. These metrics assess how well the predicted clusters align with the ground truth labels generated from manual inspections.

We compare two primary clustering methods: (1) \textit{Single-vector + kMeans}, which applies the kMeans~\cite{kMeans} algorithm to average-pooled image embeddings\footnote{We exclude results using the \texttt{[CLS]} token embedding, as they consistently underperform in our evaluations.}, and (2) \textit{Multi-vector + UPGMA~\cite{UPGMA}}, which applies hierarchical agglomerative clustering (HAC)  with `average' linkage to token-level embeddings using a symmetric $\text{MaxSim}$-based similarity measure. We use the scikit-learn clustering library~\cite{scikit-learn} to implement and evaluate clustering algorithms. Results are grouped by image ID (16 or 31) and damage category (Minor, Major, Destroyed). 

\vspace{-0.3cm}
\paragraph{Key Observations:}

\begin{itemize}[left=0pt, itemsep=6pt, topsep=6pt]
    \item \textbf{Multi-vector methods consistently outperform single-vector baselines for RemoteCLIP across most cases.} For example, in the \textit{Minor damage} category for Image ID 31, \textit{Multi-vector + UPGMA} with RemoteCLIP achieves NMI = 0.561, FMS = 0.521, and ARI = 0.416, outperforming the single-vector baseline (NMI = 0.428, FMS = 0.283, ARI = 0.177) by substantial margins. Similar trends are observed in the Major and Destroyed categories as well.
    
    \item \textbf{RemoteCLIP benefits more from multi-vector representations than SigLIP.} For Image ID 16 in the Minor category, RemoteCLIP’s multi-vector approach significantly outperforms the single-vector baseline (NMI = 0.694 vs. 0.526; FMS = 0.669 vs. 0.435;  ARI = 0.565 vs. 0.301), while SigLIP shows only modest or inconsistent gains. This suggests RemoteCLIP better encodes spatial and semantic details that can be exploited by multi-vector models.
    
    \item \textbf{SigLIP achieves the highest scores in a few settings, but the advantage is inconsistent.} For instance, in the Major category of Image ID 31, SigLIP’s single-vector method attains the best overall performance (NMI = 0.787, ARI = 0.521), slightly outperforming the multi-vector method. However, RemoteCLIP generally performs better across most other cases when paired with multi-vector embeddings.
    
    \item \textbf{Hierarchical agglomerative clustering algorithm UPGMA  with multi-vector embeddings yields better alignment with true damage clusters.} This is particularly evident in the Destroyed category for both image IDs. For example, RemoteCLIP achieves ARI = 0.393 using multi-vector UPGMA clustering for Image ID 31, compared to 0.239 with the single-vector approach.
\end{itemize}

Overall, these results demonstrate that \textbf{multi-vector embeddings combined with hierarchical agglomerative clustering algorithm (i.e., UPGMA) offer superior clustering performance}, particularly when paired with RemoteCLIP, which has been finetuned for satellite imagery~\cite{liu2024remoteclip}. While single-vector methods remain competitive in certain scenarios, especially with SigLIP (despite its lack of finetuning for satellite images), they often struggle to capture the fine-grained spatial patterns essential for accurate damage assessment. These findings underscore the value of multi-vector embeddings in strengthening post-disaster analysis pipelines.

\begin{table}[t]
    \caption{Data Ingestion Time}
    \label{table:results-ingestion-latency-summary}
    \centering
    \resizebox{\columnwidth}{!}{
        \begin{tabular}{clc}
            \toprule
            Image ID & Method & Time taken (seconds) \\
            \midrule
            31 & Baseline (w/o clustering) & 2436.1 \\
             & DamageScope & 172.9 \textit{(14.08 $\times$ faster)} \\
           \specialrule{\heavyrulewidth}{2pt}{1pt}
            \specialrule{\heavyrulewidth}{0.5pt}{3pt}
            16 & Baseline (w/o clustering) & 1335.6  \\
             & DamageScope & 168.3 \textit{(7.94 $\times$ faster)} \\
            \bottomrule
        \end{tabular}
    }
\end{table}

\subsubsection{Effectiveness of Clustering for Ingestion}
As shown in Table~\ref{table:results-ingestion-latency-summary}, DamageScope substantially reduces data ingestion latency by clustering spatially related damage instances. For Image~31, ingestion time decreases from 2{,}436.1\,s without clustering to 172.9\,s with clustering, corresponding to a \textit{$14.08\times$} speedup. Similarly, for Image~16, ingestion time decreases from 1{,}335.6\,s to 168.3\,s, yielding a \textit{$7.94\times$} speedup.

Interestingly, the baseline runtime does not scale directly with the number of property instances. Image~31 contains 139 instances, fewer than the 149 instances in Image~16, yet requires substantially more processing time (2{,}436.1\,s vs.\ 1{,}335.6\,s). The two images differ considerably in their damage composition: Image~31 contains 86 minor, 15 major, and 38 destroyed instances, whereas Image~16 contains 30 minor, 25 major, and 94 destroyed instances. This suggests that ingestion latency depends not only on the number of properties but also on characteristics of the underlying scene, such as the spatial distribution and visual complexity of the damage instances, as well as variability in model processing and API latency.

Clustering substantially reduces this instance-level workload. The 139 damage instances in Image~31 are consolidated into 20 clusters, while the 149 instances in Image~16 are reduced to 22 clusters. The key benefit of clustering is that it reduces the number of regions requiring individual processing, thereby lowering the overall ingestion workload. Consequently, DamageScope improves data ingestion efficiency and supports more scalable property-level damage analysis for disaster recovery and insurance applications.

\subsubsection{QA Performance Gains in Cost and Latency}
Our results demonstrate notable performance improvements of the DamageScope system over the baseline, particularly in terms of cost efficiency and query processing speed. As detailed in Table~\ref{table:results-inference-cost-savings-summary}, DamageScope reduces the average LLM token usage from 7{,}050 to 2{,}420 tokens per query, achieving a $2.9\times$ reduction enabled by query-aware retrieval and compact context construction. This reduction not only makes the system more cost-effective but also demonstrates that DamageScope manages the contextual information required for answering queries more efficiently.

In terms of query latency, DamageScope outperforms the baseline at every stage of the query processing pipeline. As shown in Table~\ref{table:results-inference-latency-summary}, the end-to-end QA latency decreases from 2.10 seconds to 0.69 seconds, representing an approximately $3\times$ improvement over the baseline. DamageScope achieves these latency gains by generating optimized query plans and executing them more efficiently through a dual-store data architecture.

\definecolor{DataGathering}{HTML}{D7191C}
\definecolor{QueryGeneration}{HTML}{FDAE61}
\definecolor{QueryExecution}{HTML}{ABDDA4}
\definecolor{SemanticSearch}{HTML}{2B83BA}
\definecolor{ResponseGeneration}{HTML}{926498}

\begin{table}[t]
    \caption{Question-Answering: LLM API usage cost}
    \label{table:results-inference-cost-savings-summary}
    \centering
    \begin{tabular}{lcc}
        \toprule
         && Cost (tokens) \\
        \midrule
         Baseline  && 7050 \\
         DamageScope && 2420 \textit{(2.9 $\times$ cheaper)}\\
         \bottomrule
    \end{tabular}
\end{table}

\begin{table}[t!]
    \caption{Question-Answering: Effect of Data Organization}
    \label{table:results-inference-latency-summary}
    \centering
    \begin{tabular}{lcc}
        \toprule
        && Latency (seconds) \\
        \midrule
         Baseline \textcolor{red} && 2.1 \\
         DamageScope  && 0.69 \textit{(3 $\times$ faster)} \\
         \bottomrule
    \end{tabular}
\end{table}

%% file: src/related-work.tex
\section{Related Work}

Recent advances in satellite imagery and AI have significantly enhanced post-disaster damage assessment. Early efforts used high-resolution satellite data and object-based image analysis to detect structural damage from events like earthquakes and floods~\cite{tziokas_2018_obia, zeaieanfirouzabadi_2015_flood_damage}.

Machine learning has further advanced this field. CNN-based models trained on datasets like xBD~\cite{xu_2019_building_damage_detection} have been used to classify damage, with improvements from multi-task learning~\cite{doshi2018satelliteimagerydisasterinsights} and ensemble methods~\cite{seo2019revisitingclassicalbaggingmodern}. More recent approaches extract both spectral and spatial features via dual-path networks~\cite{wang_2023_dual_path}. Multimodal techniques, such as attention-based networks using pre- and post-disaster imagery with metadata~\cite{hao_2020_attention_based}, or transformers fusing radar and optical data~\cite{chamatidis_2024_vision_transformer} have improved accuracy. Vision-language models like GeoChat enable natural language interaction with geospatial data~\cite{kuckreja2024}, while fusion frameworks combining hyperspectral and LiDAR data further enhance analysis~\cite{khodaverdi_2019_fusion_lidar_satellite}. AI-driven decision support systems now aid real-time disaster response. Tools like FEMA’s Damage Assessment Toolkit~\cite{fema2024} and automated resource planning frameworks~\cite{zhang2023} show the growing integration of generative AI in operational settings.

DamageScope builds on this foundation by combining multimodal AI, LLMs, and retrieval-augmented generation (RAG) to enable fast, cost-efficient, and interactive satellite-based damage assessment for insurers.

%% file: src/conclusion.tex
\section{Conclusion}
In this paper, we present DamageScope, a system designed to transform property damage assessment in the insurance industry by leveraging satellite imagery and AI-driven analysis. Traditional damage assessment methods rely on physical inspections, which are often costly, time-consuming, and hazardous. DamageScope addresses these challenges by automating the evaluation process using high-resolution satellite imagery, vision-language models, and the generative capabilities of large language models. By enabling rapid, large-scale assessments, DamageScope helps insurers streamline claims processing, enhance accuracy, and provide more efficient services to policyholders in the aftermath of natural disasters.

%% file: src/appendix.tex
\appendix
\section{Appendix}
\label{sec:appendix}

\subsection{DamageScope Prompt}
Listing~\ref{lst:damagescope-prompt-appendix} shows the prompt used to elicit JSON-formatted output from GPT-4o during ingestion.  

\begin{lstlisting}[style=promptstyle,caption={Extended GPT-4o prompt for structured damage extraction.},label={lst:damagescope-prompt-appendix}]
You are a damage assessment expert.
Given a satellite image of a property, analyze 
visible changes and return a JSON with the
following structure:

{
  "roof_damage": {
    "prediction": <bool>, "reason": "<why>"
  },
  "structural_damage": {
    "prediction": <bool>, "reason": "<why>"
  },
  "lawn_damage": {
    "prediction": <bool>, "reason": "<why>"
  },
  "water_damage": {
    "prediction": <bool>, "reason": "<why>"
  },
  "fire_smoke_damage": {
    "prediction": <bool>, "reason": "<why>"
  },
  "occupiability": {
    "prediction": "<habitable|
        partially_habitable|uninhabitable>", 
    "reason": "<why>"
  },
  "damage_level": {
    "prediction": <0-3>, "reason": "<summary>"
  }
}

Guidelines:
- Only describe what is visually observable.
- Avoid guessing causes.
- Provide concise reasoning sentences per field.
- Maintain consistent JSON keys and order.
\end{lstlisting}

\subsection{Example Property-Level Output}
Listing~\ref{lst:appendix-schema-example} extends the JSON schema in the main paper (Listing~\ref{lst:sample-damage-info}), showing a full record with multiple fields and reasoning strings.

\begin{lstlisting}[style=jsonstyle,caption={Extended example of structured property-level damage output.},label={lst:appendix-schema-example}]
{
  "property_id": "A12-487",
  "disaster_event": "Wildfire",
  "roof_damage": {
    "prediction": true,
    "reason": "Significant portions of the roof shingles are missing."
  },
  "structural_damage": {
    "prediction": true,
    "reason": "Walls exhibit dark scorch lines and foundation cracking."
  },
  "lawn_damage": {
    "prediction": true,
    "reason": "Vegetation fully charred and debris scattered across yard."
  },
  "water_damage": {
    "prediction": false,
    "reason": "No visible signs of water pooling or discoloration."
  },
  "fire_smoke_damage": {
    "prediction": true,
    "reason": "Heavy smoke residue observed on roof and facade surfaces."
  },
  "occupiability": {
    "prediction": "uninhabitable",
    "reason": "Structural and fire damage render the house unsafe."
  },
  "damage_level": {
    "prediction": 3,
    "reason": "Major roof loss with likely internal damage."
  }
}
\end{lstlisting}

\subsection{SQL and Vector Storage Examples}
Below we demonstrate how structured and unstructured fields from the JSON are stored separately in DamageScope’s dual-store architecture.

\begin{lstlisting}[style=jsonstyle,caption={Simplified SQL row stored in the structured store.},label={lst:appendix-sql-example}]
{
  "property_id": "A12-487",
  "roof_damage": true,
  "structural_damage": true,
  "lawn_damage": true,
  "water_damage": false,
  "fire_smoke_damage": true,
  "occupiability": "uninhabitable",
  "damage_level": 3
}
\end{lstlisting}

\begin{lstlisting}[style=jsonstyle,caption={Corresponding vector-store entry storing descriptive reasoning.},label={lst:appendix-vector-example}]
{
  "property_id": "A12-487",
  "roof_damage": "Significant portions of the roof shingles are missing.",
  "structural_damage": "Walls exhibit dark scorch lines and foundation cracking.",
  "lawn_damage": "Vegetation fully charred and debris scattered.",
  "fire_smoke_damage": "Heavy smoke residue observed on surfaces.",
  "occupiability_reason": "Structural and fire damage render the house unsafe.",
  "summary": "Major roof loss with likely internal damage."
}
\end{lstlisting}

This separation allows DamageScope to resolve both factual and explanatory queries efficiently, i.e. SQL queries for numeric/categorical attributes, vector queries for semantic reasoning.